\pdfoutput=1
\documentclass[11pt]{article}

\usepackage[]{acl}

\usepackage{times}
\usepackage{latexsym}

\usepackage[T1]{fontenc}
\usepackage[utf8]{inputenc}

\usepackage{microtype}

\usepackage{booktabs, colortbl} 
\usepackage{multirow,makecell} 
\usepackage{multicol}
\usepackage{amsmath}
\usepackage{amssymb}
\usepackage{caption}
\usepackage{subcaption}
\usepackage{titlesec}
\usepackage{pgfplots}
\usepackage[normalem]{ulem} 
\usepackage{enumitem}
\usepackage{graphicx} 
\usepackage{hyperref}
\usepackage{tabularx}
\usepackage{CJK}

\colorlet{LightGray}{gray!40}

\title{KIT's Multilingual Speech Translation System for IWSLT 2023}

\author{Danni Liu, Thai Binh Nguyen, Sai Koneru, Enes Yavuz Ugan, Ngoc-Quan Pham,\\ 
\textbf{Tuan-Nam Nguyen, Tu Anh Dinh, Carlos Mullov, Alexander Waibel, Jan Niehues} \\
  Karlsruhe Institute of Technology \\
  \texttt{firstname.lastname@kit.edu} 
}
\begin{document}
\maketitle
\begin{abstract}
Many existing speech translation benchmarks focus on native-English speech in high-quality recording conditions, which often do not match the conditions in real-life use-cases.
In this paper, we describe our speech translation system for the multilingual track of IWSLT 2023, which evaluates translation quality on scientific conference talks.
The test condition features accented input speech and terminology-dense contents.
The task requires translation into 10 languages of varying amounts of resources.
In absence of training data from the target domain, we use a retrieval-based approach ($k$NN-MT) for effective adaptation ($+0.8$ BLEU for speech translation).
We also use adapters to easily integrate incremental training data from data augmentation, and show that it matches the performance of re-training.
We observe that cascaded systems are more easily adaptable towards specific target domains, due to their separate modules.
Our cascaded speech system substantially outperforms its end-to-end counterpart on scientific talk translation, although their performance remains similar on TED talks.\footnote{Code available at \url{https://github.com/dannigt/kit-iwslt2023-multilingual}}
\end{abstract}

\section{Introduction}

This paper summarizes Karlsruhe Institute of Technology's  speech translation system for the multilingual track of IWSLT 2023 \cite{iwslt:2023}.
In this track, the task is to translate scientific talks in English into 10 languages: Arabic (ar), Chinese (zh), Dutch (nl), French (fr), German (de), Japanese (ja), Persian/Farsi (fa), Portuguese (pt), Russian (ru), Turkish (tr).
The talks are from presentations in the 60th Annual Meeting of the Association for Computational Linguistics (ACL 2022) and compiled by \citet{salesky-2023-evaluating}.

Translating scientific talks presents  several challenges.
On the \textit{source} side, most speakers are non-native, and the recording conditions often vary.
This requires acoustic robustness to accents and noise.
On the \textit{target} side, domain-specific terminologies are frequently used, calling for accurate translation of these words that rarely occur in the training data.
The styles of the talks, e.g. formality, also differ from other domains.
As no training data from the same domain is provided,
effective few-shot or zero-shot adaptation is crucial.

As the task focuses on \textit{one-to-many} translation, it is also an interesting testbed for whether multilinguality improves  speech translation quality.
For text-to-text translation, the gain from multilinguality is mostly concentrated in \textit{many-to-one} translation~\cite{aharoni-etal-2019-massively, mm100}, i.e., multilinguality on the source side. 
In contrast, for \textit{X-to-many} translation, it remains unclear whether incorporating \textit{more target languages} improves translation quality.

In this system description paper, 
we present cascaded and end-to-end systems for the English-to-many speech translation task.
We leverage pretrained models, including WavLM~\cite{wavlm}, mBART50~\cite{mbart50}, and DeltaLM~\cite{deltalm}.
The systems do not use additional data beyond the allowed corpora, and therefore fall under the \textit{constrained} data condition.

For the cascaded system, 
to handle the unique style of scientific talks, we use $k$NN-MT~\cite{knnmt} to bias the output generation towards the target domain.
Moreover, as no target monolingual data is provided, we use data diversification~\cite{data_div} to enrich the existing parallel data.
We also use adapters~\cite{adapters_orig,bapna-firat-2019-simple} as a lightweight approach for incremental learning and language adaptation.
For the ASR model, we improve over last year's performance by using a more recent audio encoder~\cite{wavlm} and adding a dedicated decoder.
To adapt the ASR system to the target domain, we use $n$-gram re-weighting and synthesized data for the target domain.
For the end-to-end system, we use our machine translation model for knowledge distillation.
We also ensemble models trained with and without synthesized speech data.

Our main findings are as follow:
\begin{itemize}
\item For cascaded ST systems, we can effectively adapt the model towards a target domain/style using $k$NN-MT~\cite{knnmt}. 
A datastore as small as a few hundred sentence pairs was sufficient for achieving consistent gains (avg. $+0.8$ BLEU over 10 languages).
\item Besides the common use-case of adding language-specific capacity, adapters~\cite{bapna-firat-2019-simple} is also an effective method when subsequently adding training data. Empirically, we show it matches the performance of re-training on all new data.
\item For ASR, lexical constraints for domain adapation are more easily integrated in CTC models.
For encoder-decoder model, the control could be achieved by TTS-synthesized source speech, but it requires more careful tuning.
\end{itemize}
    
% TODO:
% \begin{itemize}
%     \item describe task
%     \item describe challenge
%     \item overall approach
%     \item describe main novelties
%     \item describe main findings
% \end{itemize}

\section{Data and Preprocessing}
After describing the evaluation data (\S\ref{subsec:eval_data}), 
we outline the training data and preprocessing steps for our automatic speech recognition (ASR; \S\ref{subsec:asr_data}), machine translation (MT; \S\ref{subsec:mt_data}), casing/punctuation restoration (\S\ref{subsec:punc_data}), and speech translation (ST; \S\ref{subsec:st_data}) models. 

\subsection{Development and Test Data} \label{subsec:eval_data}
In the multilingual track, 
the testing condition is scientific conference talks.
Therefore, we primarily rely on the ACL development (dev) set \cite{salesky-2023-evaluating} for validation.
It consists of English transcripts of the talks and translations into the $10$ target languages.
The systems are then evaluated on a blind test set.
The dev and test sets consist of $5$ talks each.
The paper abstracts for all talks are available in English.
The talks are pre-segmented. In all experiments, we use the given segmentation.

We also report performance on tst-COMMON of MuST-C~\cite{di-gangi-etal-2019-must}, tst2019 and tst2020 from previous years' evaluations~\cite{anastasopoulos-etal-2021-findings, anastasopoulos-etal-2022-findings}.

An overview of the development and test data is in \autoref{tab:dev_test_data}.

\begin{table}[ht!]
\small
\centering
\setlength\tabcolsep{2pt} % default value: 6pt
\begin{tabular}{lrrr}
\toprule
\textbf{Dev/Test set} & 
\textbf{Hours} &
\textbf{\# Utterances} & 
\textbf{Domain} \\ 
\midrule
ACL dev      & 1.0  & 468 & ACL conference talks \\
tst-COMMON    & 4.9 & 2823 & TED talks \\
tst2019      & 4.8 & 2279 & TED talks \\
tst2020      & 4.1 & 1804  & TED talks \\
\bottomrule
\end{tabular}
\caption{\label{tab:dev_test_data} Overview of development and test data.}
\end{table}

\begin{table}[ht!]
\small
\centering
\setlength\tabcolsep{10pt} % default value: 6pt
\begin{tabular}{lrr}
\toprule
\textbf{Corpus / Data Source} & \textbf{Hours} & \textbf{\# Utterances}  \\ 
\midrule
Common Voice    & 1667 & 1225k \\
LibriSpeech     & 963 & 281k \\
MuST-C v2       & 482 & 251k \\
TED-LIUM v3     & 452 & 268k \\
VoxPopuli       & 501 & 177k \\
\midrule
TTS             & 7284 & 4.7M \\
\bottomrule
\end{tabular}
\caption{\label{tab:asr_data} ASR data overview.}
\end{table}

\subsection{Speech Recognition Data} \label{subsec:asr_data}

For the ASR training, we use Common Voice~\cite{cv}, LibriSpeech~\cite{librispeech}, MuST-C v2~\cite{di-gangi-etal-2019-must}, TED-LIUM v3~\cite{tedlium3}, and VoxPopuli~\cite{wang-etal-2021-voxpopuli}.
The data overview is in~\autoref{tab:asr_data}.

\paragraph{Synthesized Speech Data} 
To adapt the ASR model to the ACL talks, we add synthesized speech created by a text-to-speech (TTS) model.
Specifically, from the MT bitext English side  (\autoref{tab:mt_data}), we select sentences similar to the ACL domain based on similarity with the provided ACL dev bitext and abstracts.
Inspired by data selection strategies for MT \cite{eck-etal-2005-low,DBLP:journals/corr/abs-2201-05700}, we use $n$-gram overlap as similarity metric. 
4.7M sentences are selected and then synthesized to speech by a VITS~\cite{vits} model trained on MuST-C.
The synthesized data amount is shown in the last row of \autoref{tab:asr_data}.

\subsection{Machine Translation Data} \label{subsec:mt_data}

The MT training data include the following text-to-text translation corpora: Europarl v7 and v10~\cite{koehn-2005-europarl}, NewsCommentary v16, OpenSubtitles v2018~\cite{lison-tiedemann-2016-opensubtitles2016}, Tatoeba \cite{tiedemann-2012-parallel}, and ELRC-CORDIS\_News, JParaCrawl~\cite{morishita-etal-2022-jparacrawl} for Japanese, and TED2020~\cite{reimers-gurevych-2020-making} for German\footnote{This dataset has deplication with past evaluation sets: tst2019 tst2020 and tst-COMMON. The deplications were removed prior to training.}.
We also include the text translation part of the following ST corpora: MuST-C~\cite{di-gangi-etal-2019-must}, CoVoST v2~\cite{covost}, and Europarl-ST~\cite{europarlst}. 
The aggregated data amount per language is summarized in the ``Original'' column of \autoref{tab:mt_data}.

\begin{table}[h]
\small
\centering
\setlength\tabcolsep{6.5pt} % default value: 6pt
\begin{tabular}{lrrrrr}
\toprule
& \textbf{Original} 
&& \multicolumn{2}{c}{\textbf{After Diversification}} 
\\
\cmidrule{2-2} \cmidrule{4-5}
\textbf{Lang.} 
& \textbf{\# sent. (M)}  
&& \textbf{\# sent. (M)} 
& \textbf{\# tokens (M)} \\ 
\midrule
ar      & 26.0   && 65.2 & 865.0\\
zh      & 11.2   && 21.5 & 254.3 \\
nl      & 33.1   && 82.1 & 1162.7\\
fr      & 38.9   && 91.6 & 1427.8\\
de      & 23.0   && 54.4 & 860.0 \\
ja*      & 2.6    && 27.2 & 832.7\\
fa      & 5.8    && 11.3 & 162.1 \\
pt      & 29.0   && 72.3 & 1024.3\\
ru      & 22.1   && 51.5 & 685.3\\
tr      & 36.7   && 89.7 & 1021.2\\
\midrule
Total   & 228.4 && 566.8 & 8295.4\\
\bottomrule
\end{tabular}
\caption{\label{tab:mt_data} MT data overview.
*: For ja, the original data of 2.6M sentences did not include JParaCrawl, which was announced later as allowed data.}
\end{table}

As preprocessing, we perform truecasing, deduplication, length ratio filtering, and histogram filtering using the statistics by~\citet{mm100}.
Then we perform subword segmentation using Sentencepiece~\cite{kudo-richardson-2018-sentencepiece} based on the vocabulary of mBART50~\cite{mbart50}.

\paragraph{Data Diversification}
Different from last years' shared tasks~\cite{anastasopoulos-etal-2021-findings, anastasopoulos-etal-2022-findings}, no monolingual (non-English) data is provided.
This means conventional data augmentation techniques like backward translation are not directly applicable.
On the other hand, forward translation from existing English monolingual data may introduce undesirable errors in the translation targets, especially on lower-resource languages. 
In this light, we use data diversification~\cite{data_div}, a data augmentation method that enriches existing parallel data by forward and backward translating the training bitext.
As the model has seen the parallel data in training, the synthetic translations are expected to have relatively high quality.
Moreover, either the source or target side of the synthetic data is from the original bitext.
The diversified data amount after deduplication is shown in \autoref{tab:mt_data}.
Here we perform one round of forward and backward translation, as ~\citet{data_div} have empirically shown further rounds do not lead to substantial gains.

\subsection{Casing/Punctuation Restoration Data} \label{subsec:punc_data}
The ASR outputs are lower-cased and unpunctuated, while the MT model expects cased and punctuated inputs.
We randomly sample 1.5 million English sentences from the MT training data (\autoref{tab:mt_data}), and remove the casing and punctuation marks as training source data.
We then train a model to restore the casing and punctuation marks.

\subsection{Speech Translation Data} \label{subsec:st_data}
The speech translation data are shown in \autoref{tab:st_data}.
We additionally use our trained MT model to create forward translations based on the following transcript-only datasets: Common Voice, TEDLIUM, and VoxPopuli.
The TTS data described in \S\ref{subsec:asr_data} is also used.

\begin{table}[h]
\small
\centering
\setlength\tabcolsep{5.5pt} % default value: 6pt
\begin{tabular}{llrr}
\toprule
\textbf{Lang.} 
& \textbf{Corpus / Data Source} 
& \textbf{Hours}
& \textbf{\# Utterances}   \\
\midrule
ar         & CoVoST & 429 & 289k \\
           & MuST-C & 463 & 212k \\
           & TTS & 283 & 203k \\
zh         & CoVoST & 429 & 289k \\
           & MuST-C & 596 & 358k  \\ %check seg duration 
           & TTS & 204 & 183k \\
nl         & MuST-C & 434 & 248k \\
           & europarl-ST & 75 & 32k \\
           & TTS & 1138 & 713k \\
fr         & MuST-C & 485 & 275k \\
           & europarl-ST & 76 & 32k \\
           & TTS & 1768 & 998k \\
de         & CoVoST & 429 & 289k \\
           & MuST-C & 440 & 269k \\
           & europarl-ST & 77 & 33k \\
           & TTS & 1891 & 779k \\
ja         & CoVoST & 429 & 289k \\
           & MuST-C & 541 & 329k \\
           & TTS & 73 & 56k \\
fa         & CoVoST & 429 & 289k \\
           & MuST-C & 347 & 182k \\
           & TTS & 89 & 88k \\ 
pt         & MuST-C & 377 & 206k \\
           & europarl-ST & 75 & 32k \\
           & TTS & 1678 & 639k \\
ru         & MuST-C & 482 & 265k\\
           & TTS & 331 & 331k \\
tr         & CoVoST & 429 & 289k \\
           & MuST-C & 446 & 236k\\
           & TTS & 428 & 511k \\
\midrule
all        & Common Voice    & 1488  & 948k \\
           & TEDLIUM    & 453 & 268k \\
           & VoxPopuli  & 502 & 177k \\
\bottomrule
\end{tabular}
\caption{\label{tab:st_data} ST data overview.
The last section ``all'' indicates forward translated synthetic targets from transcript-only corpora, which are available for all 10 languages.}
\end{table}

\section{Cascaded System}
For the cascaded system, we introduce our ASR (\S\ref{subsec:res_asr}) and MT (\S\ref{subsec:res_mt}) models.

\subsection{Automatic Speech Recognition Module} \label{subsec:res_asr}
\paragraph{Baseline Models}
The first baseline is our ASR model for last year's offline track~\cite{pham-etal-2022-effective}.
It is a Wav2vec 2.0~\cite{w2v2} with \textsc{Large} configuration pretrained on $960$ hours of Librispeech data.
This year, after seeing initial favourable results compared to Wav2vec, we opt for WavLM~\cite{wavlm} as audio encoder.
We use the \textsc{Large} configuration with $24$ layers.
We use the mBART50~\cite{mbart50} decoder along with the WavLM encoder.
As the ASR model only needs to transcribe English\footnote{BART, the English-only predecessor of mBART, is not among the allowed pretrained models.}, 
we trim the mBART50 vocabulary from $256$k down to $62$k tokens by removing all non-alphabetic tokens.

\paragraph{In-Domain TTS Data}
We also use the synthesized TTS data. 
Compared to the same model without TTS data, the word error rate (WER) improves from $11.6\%$ to $10.7\%$ on ACL dev, but degrades from $8.4\%$ to $9.0\%$ on the TEDLIUM test set.
There are two potential explanations: 
First, the noisy TTS speech may be helpful for handling the non-native utterances prominent in the ACL dev set.
Second, the target side of the TTS data is more relevant to the ACL domain, as we selected them based on $n$-gram overlap with ACL data. 
This in turn improves ASR performance on the ACL dev set.

As shown in \autoref{tab:asr_res}, compared to last year's submission, this year's ASR model achieves consistent gains across domains on ACL dev, tst-COMMON, and tst2020.

\begin{table}[ht!]
\small
\centering
\setlength\tabcolsep{2pt} % default value: 6pt
\begin{tabular}{lcccccccc}
\toprule
\textbf{Model} 
& \textbf{ACL dev} 
& \textbf{tstCom.} 
& \textbf{tst2020} \\  % add after TTS
\midrule
ASR 2022~\cite{pham-etal-2022-effective} 
& 12.5 & 5.4 & 5.6 \\
WavLM + mBART50 & 10.7 & 3.9 & 4.8 \\
\bottomrule
\end{tabular}
\caption{\label{tab:asr_res} ASR results in WER($\downarrow$) in comparison to our submission last year~\cite{pham-etal-2022-effective} which used Wav2vec trained with CTC and a $5$-gram LM.
By using WavLM audio encoder and the mBART decoder, we achieve consistent gains across domains (ACL and TED, i.e., tst*).}
\end{table}

\paragraph{Language Model (LM) Adaptation}
Aside from using TTS data, we also investigate other methods to adapt towards the ACL domain using the provided paper abstracts.
On preliminary experiments with Connectionist Temporal Classification (CTC) + $n$-gram LM models, we integrate ACL abstract $5$-grams statistics into the language models.
As shown in the upper section of \autoref{tab:asr_res_adapt}, this improves on ACL dev (WER $13.8\%\rightarrow13.0\%$) while preserving the performance on TED talks (tst-COMMON WER stays at $7.6\%$).

As our final system is an encoder-decoder model (WavLM + mBART50), adapting the LM alone is less straightforward.
We create pseudo ASR training data with ACL data on the transcript side. 
Specifically, we use our TTS model to synthesize speech from the ACL dev and test abstracts.
As the amount of ACL abstract data is very limited (less than $100$ sentences in total), we heavily upsampled them, so that they consist of $60\%$ of the training data.
As shown in the lower section of \autoref{tab:asr_res_adapt}, 
this leads to a minor improvement of WER for ACL dev.
However, the gain does not carry over to ST performance when later cascading with our MT model.
Therefore, our final ASR system did not use the abstracts.
The lack of improvement could be related to the low amount of ACL abstract data, which requires heavy upsampling of the TTS data, 
and as a result hinders the ability of transcribing real speech.

The contrast between the two sets of experiments may be related to diminishing gains as WER improves, i.e., for the Wav2vec + CTC + LM model, gaining over a WER of $13.8\%$ is easier than starting from a $10.7\%$ WER.
Another interpretation of the difference could be that adding specific constraints to ``end-to-end'' ASR models is more challenging than the counterparts with separate LMs.

\begin{table}[ht!]
\small
\centering
\setlength\tabcolsep{1.5pt} % default value: 6pt
\begin{tabular}{lcccccccc}
\toprule
\textbf{Model} 
& \textbf{ACL dev} 
& \textbf{tst-COMMON} \\  % add after TTS
\midrule
Wav2vec + CTC + $5$-gram & 13.8 & 7.6 \\
+ ACL abstract $5$-gram  & 13.0 & 7.6 \\
\midrule
WavLM + mBART50 & 10.7 & 3.9 \\
+ ACL abstract TTS (upsampled) & 10.5 & 4.3 \\
\bottomrule
\end{tabular}
\caption{\label{tab:asr_res_adapt} ASR adaptation results in WER($\downarrow$). On preliminary experiments with Wav2vec + CTC + LM models, we improve ASR performance on ACL dev by integrating $n$-gram statistics from the ACL abstracts.
For the WavLM + mBART 50 model, adding synthesized audio-transcript data based ACL dev abstracts does not give consistent gain.}
\end{table}

\begin{table*}[h]
\small
\centering
\setlength\tabcolsep{3.5pt} % default value: 6pt
\begin{tabular}{lcccccccccccccccc}
\toprule
& \multicolumn{11}{c}{\textbf{ACL dev (en$\rightarrow$X)}} 
&& \multicolumn{2}{c}{\textbf{TED (en$\rightarrow$de)}}  \\
\cmidrule{2-12} \cmidrule{14-15} 
\textbf{ID} & \textbf{de} & \textbf{ja} & \textbf{zh}  & \textbf{ar} & \textbf{nl} & \textbf{fr} & \textbf{fa} & \textbf{pt} & \textbf{ru} & \textbf{tr} 
&\cellcolor[HTML]{d3d1d1}\textbf{Avg.}   
&& \textbf{tst2019} & \textbf{tst2020} \\ 
\midrule
\multicolumn{5}{l}{From ground-truth transcripts (\textbf{MT alone})} \\
$(1)$ base 
& 39.8 
& 44.2
& 47.4 
& 30.4 
& 45.7
& 48.9
& 23.6
& 51.1 
& 19.5
& 22.9 
&\cellcolor[HTML]{d3d1d1}37.4 
&& 29.5
& 32.9 \\
$(2)$ data divers. all
& 41.6 
& 44.5
& 49.8
& 33.6
& 50.7
& 51.1
& 25.4
& 52.5 
& 21.5
& 24.6 
&\cellcolor[HTML]{d3d1d1}39.5 
&& 30.0
& 33.7 \\
$(3)$ $(1)$ + data divers.; adapter 
& 41.4 
& 45.8
& 48.8
& 33.3
& 49.8
& 51.5
& 25.2
& 54.1
& 21.9
& 24.1
&\cellcolor[HTML]{d3d1d1}39.6
&& 29.5
& 33.2 \\
$(4)$ ensemble $(2)$ + $(3)$
& 41.7 
& 46.1
& 49.6
& 33.7
& 50.8
& 52.1
& 25.9
& 54.3
& 23.1
& 24.8
&\cellcolor[HTML]{d3d1d1}40.2 
&& 30.4
& 33.7 \\
$(5)$ $(4)+k$NN-MT
& 43.7 
& 47.3
& 49.8
& 35.4
& 52.3
& 52.8
& 27.2
& 55.3
& 23.9
& 27.1
&\cellcolor[HTML]{d3d1d1}41.5 
&& 30.4 
& 33.4\\
\midrule
\multicolumn{3}{l}{From ASR outputs (\textbf{cascaded ST})} \\
$(1)$ base
& 34.3 
& 38.2
& 41.6
& 25.3
& 36.6
& 39.9
& 19.1
& 40.7
& 16.7
& 18.9
&\cellcolor[HTML]{d3d1d1}31.1 
&& 26.5
& 28.0 \\
$(2)$ data divers. all
& 35.4
& 38.6
& 44.3
& 26.8
& 39.2
& 41.5
& 20.5
& 42.6
& 18.7
& 19.5
&\cellcolor[HTML]{d3d1d1}32.7
&& 27.0
& 29.3
\\
$(3)$ $(1)$ + data divers.; adapter 
& 35.5
& 39.0
& 43.6
& 26.4
& 38.9
& 41.9
& 20.2
& 43.0
& 19.3
& 19.6 
&\cellcolor[HTML]{d3d1d1}32.7 
&& 26.7
& 28.3
\\
$(4)$ ensemble $(2)$ + $(3)$
& 36.1
& 39.8
& 44.4
& 26.9
& 39.8
& 42.3
& 20.7
& 43.5
& 19.2
& 19.7 
&\cellcolor[HTML]{d3d1d1}33.2 
&& 26.9
& 28.7 \\
$(5)$ $(4)$ + $k$NN-MT
& 36.8
& 40.2
& 44.6 
& 28.2 
& 40.8
& 42.0
& 21.8
& 44.5 
& 19.7 
& 21.1 
&\cellcolor[HTML]{d3d1d1}34.0 
&& 26.9
& 28.5 \\
\midrule
\multicolumn{3}{l}{\textbf{End-to-end ST}} \\
$(6)$ WavLM + mBART50 decoder
& 31.7
& 29.2
& 40.7
& 25.0
& 36.7
& 40.5
& 19.5
& 43.0
& 16.9
& 18.5
&\cellcolor[HTML]{d3d1d1}30.2 
&& 27.0
& 29.3 \\
$(7)$ $(6)$ + TTS
& 33.2
& 29.2
& 40.5
& 25.5
& 37.9
& 41.0
& 20.1
& 43.9
& 16.5
& 18.9
&\cellcolor[HTML]{d3d1d1}30.7
&& 27.0
& 29.1
\\
$(8)$ ensemble $(6)$ + $(7)$
& 34.0
& 29.9
& 41.7
& 25.5
& 38.2
& 42.0
& 20.2
& 44.4
& 18.3
& 20.2 
&\cellcolor[HTML]{d3d1d1}31.4 
&&  27.3
& 29.6
\\
\bottomrule
\end{tabular}
\caption{\label{tab:st_res} MT and ST results in BLEU$(\uparrow)$.
We evaluated against the original references without NFKC normalization, although the hypotheses are normalized by the default ``\textsc{nmt\_nfkc}'' rule in SentencePiece \cite{kudo-richardson-2018-sentencepiece}.}
\end{table*}

\paragraph{Casing/Punctuation Restoration}
We take a sequence-to-sequence approach to the casing and punctuation restoration problem.
Specifically, we train a punctuation model initializing from DeltaLM-base~\cite{deltalm} to restore the casing and punctuation information, using the training data described in \S\ref{subsec:punc_data}.

\subsection{Machine Translation Module} \label{subsec:res_mt}
\paragraph{Baseline Model}
We start with the pretrained DeltaLM~\cite{deltalm} with \textsc{Large} configuration.
The pretrained model has $24$ and $12$ encoder and decoder Transformer layers respectively. 
It uses postnorm layer normalization.
It is a fully multilingual model where all parameters are shared across languages.
The target language tokens are prepended to the source target sentences.
We use temperature-based sampling~\cite{arivazhagan2019massively}
with $\tau=5.0$ to counteract the data imbalance between languages.
When training, we use a relatively large effective batch size of $128$k as preliminary experiments with smaller batch sizes showed more instabilities in training. 
This might be a side effect of the postnorm layer normalization~\cite{prenorm}.
The results of the baseline are shown in Row $(1)$ of \autoref{tab:st_res}, with an average score of $37.4$ BLEU\footnote{By default using \texttt{tok.13a} from sacreBLEU \cite{post-2018-call}, except for zh and ja where we use \texttt{tok.zh} and \texttt{tok.ja-mecab-0.996-IPA}.} on ACL dev.

\paragraph{Data Diversification}
As motivated in \S\ref{subsec:mt_data}, 
we use data diversification as an alternative data augmentation method in absence of monolingual target data for backtranslation.
As data diversification needs forward and backward translations on the training data,
we additionally train a $10$-to-English model to create the backward translations.
Row $(2)$ of \autoref{tab:st_res} shows the results after data diversification on all languages pairs.
On average, this data augmentation approach improves MT quality by $2.1$ BLEU and ($37.4\rightarrow39.5$), and ST quality by $1.6$ BLEU ($31.1\rightarrow32.7$).

\paragraph{Adapters for Incremental Data}
Retraining on the new training data after diversification (Row $(2)$ of \autoref{tab:st_res}) is time-consuming and costly.
To adapt the initial model (Row $(1)$ of \autoref{tab:st_res}) rapidly towards to the augmented data, we use adapters~\cite{bapna-firat-2019-simple,philip-etal-2020-monolingual}.
In this case, the adapters are target-language-specific.
The adapters are inserted after each encoder and decoder layer. 
We initialize from the trained baseline (Row $(1)$ in \autoref{tab:st_res}),
freeze trained parameters and update the adapters only.
We use the efficient implementation from~\citet{baziotis-etal-2022-multilingual}.
As shown in Row $(3)$ of \autoref{tab:st_res}, only training the adapters on the new diversified training data performs on par with the re-training setup in Row (2) ($39.6$ on MT and $32.7$ on ST on average for ACL dev).
These results demonstrate that adapters are suitable for fast and effective incremental learning when additional training data emerges later.

To our surprise, adding adapters to the model trained with full data diversification (Row $(2)$ from \autoref{tab:st_res}) does not bring further gain.
%We were originally hoping to see improvements on high-resource languages, which supposedly faced capacity bottleneck.
A similar observation was reported by \citet{DBLP:journals/corr/abs-2305-02665}, who opted for training the full network from scratch along with adapters instead.
In our case, it therefore would be interesting to see the impact of training on data diversification with adapters from scratch.

\paragraph{Multilingual vs Bilingual}
To investigate the impact of interference from multiple target languages, 
in preliminary experiments, we also compare the multilingual and bilingual translation performance for selected language pairs.
As shown in \autoref{tab:mt_adapters},
compared to bilingual models, the multilingual model lags behind especially on higher-resource languages.
Adding the adapters partly closes this gap.
Note the score difference to main result table (\autoref{tab:st_res}) is because the preliminary experiments did not fully use diversified data for all languages.

\begin{table}[ht!]
\small
\centering
\setlength\tabcolsep{3pt} % default value: 6pt
\begin{tabular}{lrrrrrrr}
\toprule
\textbf{Model} &
\multicolumn{3}{c}{\textbf{ACL dev}} 
&& \multicolumn{3}{c}{\textbf{tst-COMMON}} \\
\cmidrule{2-4} \cmidrule{6-8} 
& \textbf{en-de} & \textbf{en-ru} & \textbf{en-fa} && \textbf{en-de} & \textbf{en-ru} & \textbf{en-fa} \\
\midrule
bilingual   
& 41.0 & 20.0 & 24.2  
&& 34.3 & 22.7 & 16.0   \\
multilingual 
& 39.8 & 19.5 & 23.6  
&& 34.1 & 21.9 & 15.9  \\
+ adapters 
& 40.9 & 20.2 & 23.7  
&& 34.7 & 22.2 & 16.3  \\
\bottomrule
\end{tabular}
\caption{\label{tab:mt_adapters} 
Comparison of bilingual vs multilingual translation performance in BLEU $(\uparrow)$ on German (de), Russian (ru), Farsi (fa), which are high-, mid-, low-resource in the training data (\autoref{tab:mt_data}).
Multilingual system falls behind bilingual system, while adapters partly closes the gap.
Note the score difference to main result table (\autoref{tab:st_res}) is because the experiments here did not fully use diversification.}
\end{table}

\paragraph{Ensemble}
Although the models in Row $(2)$ and $(3)$ in \autoref{tab:st_res} are trained on the same data and share the same base architecture, we expect their representations to be sufficiently different, as $(3)$ additionally uses adapters.
We therefore ensemble these two models.
The results are in Row $(4)$ of \autoref{tab:st_res}.
On MT and ST, for ACL, ensembling shows an improvement of $0.6$ and $0.5$ BLEU respectively over the single models in Row $(2)$ and $(3)$.
On TED, however, ensembling does not seem to impact the scores compared to the single models.
One explanation is that the adapter model from Row $(3)$ performs worse than its non-adapter counterpart (Row $(2)$) on TED, which limits the overall effectiveness of ensembling.
%The difference may be related to  domains, where the ACL set has a larger domain shift from the training data, and therefore benefits more from ensembing.

\begin{table}[h]
	\centering
	\small 
	\begin{tabularx}{\columnwidth}{X}
	\toprule
		\textbf{Source} (ASR output): ... in a {\color{brown}zero shot evaluation setup}, meaning that {\color{brown}pre trained} word embedding models are applied out of the box without any additional fine tuning
            \\
		\textbf{w/o $k$NN-MT} (\autoref{tab:st_res} row $(4)$): ... in einer {\color{purple}Null-Shot-Bewertungs-Setup} {\color{gray}(zero-shot evaluation setup)}, was bedeutet, dass {\color{purple}vorgebildete} {\color{gray}(pre-educated)} Wort-Einbettungsmodelle ohne zus\"atzliche Feinabstimmung direkt angewendet werden. 
		\\
            \textbf{w/ $k$NN-MT} (\autoref{tab:st_res} row $(5)$): ... in einer {\color{blue}Null-Shot-Bewertung} {\color{gray}(zero-shot evaluation)}, was bedeutet, dass {\color{blue}vortrainierte} {\color{gray}(pretrained)} Wort-Einbettungsmodelle ohne zus\"atzliche Feinabstimmung direkt angewendet werden. \\
        \midrule
            \textbf{Source} (ASR output): {\color{brown}Hello}. My name is Ramachandra, and I will {\color{brown}present} our paper. \\
            \textbf{w/o $k$NN-MT} (\autoref{tab:st_res} row $(4)$): \begin{CJK*}{UTF8}{gbsn}{\color{purple}你好} {\color{gray}(Hello; addressing a single person)},我叫拉玛钱德拉\hspace{4pt}我要{\color{purple}发表} {\color{gray}(publish)}我们的论文\end{CJK*}\\
            \textbf{w/ $k$NN-MT} (\autoref{tab:st_res} row $(5)$): \begin{CJK*}{UTF8}{gbsn}{\color{blue}大家好} {\color{gray}(Hi all; addressing a group of audience)},我叫拉玛钱德拉, 我要{\color{blue}介绍} {\color{gray}(introduce)}我们的论文。\end{CJK*}\\
	\bottomrule
	\end{tabularx}	
	\caption{Examples of $k$NN-MT improving translation quality for en$\rightarrow$de (upper) and en$\rightarrow$zh (lower). 
    $k$NN-MT creates more accurate terminology translations (``pre trained'' for en$\rightarrow$de) and create more context-appropriate translation (``Hello'' for en$\rightarrow$zh).}\label{tab:example_knn}
\end{table} 

\paragraph{$k$NN-MT}
We also adapt the MT model to the target domain of scientific talks. 
A challenge is that we do not have sufficient training data to fully finetune the MT model towards the desired domain or style.
In this case, we use $k$NN-MT~\cite{knnmt} to adapt the model at inference time.
In $k$NN-MT, bitexts are passed through a trained MT model. 
For each target token, its decoder hidden state is stored in a datastore.
At inference time, based on the current decoder hidden state, $k$ candidate target tokens are retrieved from the datastore using a nearest neighbor lookup.
The retrieved token distribution is then interpolated with the MT target distribution, which in turn generates the output tokens.
Hyperparameters for $k$NN-MT include the number of retrieved neighbors $k$,
the temperature for smoothing the $k$NN distribution $T$, and the interpolation weight $w$.

In our experiments, we use systems $(2)$ and $(3)$ from \autoref{tab:st_res} for creating the datastores.
As different models' hidden states (which serve as keys in the datastore) also differ substantially, the datastore is MT-model-dependent.
To use $k$NN-MT when ensembling systems $(2)$ and $(3)$, we therefore need two datastores for systems $(2)$ and $(3)$ respectively.
The $k$NN-MT candidate tokens are interpolated with the output vocabulary distribtuion before the ensembling operation.

We use hyperparameters $k=8$, $T=50$, $w=0.3$, after an initial search with $T \in [10, 50, 100], w \in [0.1, 0.3, 0.5]$.
Our implementation mostly follows ~\citet{zheng-etal-2021-adaptive}, which uses the \textsc{Faiss} toolkit \cite{johnson2019billion} for efficient $k$NN operations.
Comparing the inference speed of system $(4)$ and $(5)$, 
with the same batch size of $64$ sentences\footnote{System $(5)$ requires more GPU memory than system $(4)$. The latter would be able to use a larger batch size of 128 sentences, making the realistic speed difference slightly larger.}, 
using $k$NN-MT takes roughly $50\%$ more time on a Nvidia Titan RTX GPU with 24GB memory.

Naively using all ACL dev bitext as datastore would lead the model to copying the oracle targets.
To simulate the scenario on the blind test set, 
when translating the $i$-th talk, we use the other $j_{j{\neq}i}\in[n]$ talks' bitext as datastore, where $n$ is the total number of talks.

As shown in Row $(5)$ of \autoref{tab:st_res}, $k$NN-MT brings an additional gain of $1.3$ BLEU on MT and $0.8$ BLEU on ST.
These results shows a datastore as small as hundreds of sentence pairs can be effectively used for inference-time domain adaptation.

\autoref{tab:example_knn} shows two examples of $k$NN-MT improving translation quality, apart from generic improvements in fluency and accuracy, in these examples $k$NN-MT also helps generate correct terminologies and context-appropriate greetings.

%In the final submission, we also add the translation of the abstracts to the datastore.

\section{End-to-End System}
For the end-to-end system, 
similar to our ASR model,
after seeing initial favourable results of WavLM over Wav2vec, we choose WavLM as the audio encoder.
Following last year's submission~\cite{pham-etal-2022-effective}, we use the mBART50 decoder.
The results are shown in Row $(6)$ of \autoref{tab:st_res}.
Contrasting Row $(6)$ and $(7)$ reveals that adding the TTS data does not substantially change ST performance. 
However, ensembling the two models trained with and without TTS data (Row $(8)$) improves over the single models (on average $+0.7$ for ACL, $+0.4$ for TED), despite them having the identical architecture.

Compared to the strongest cascaded system (Row $(5)$), the end-to-end system falls behind $2.6$ BLEU on ACL dev.
On TED, however, it appears to slightly outperform the cascaded system.
One explanation is that the MT model of the cascaded system has not been separately adapted to TED texts (although parts of the full training data do cover TED data), which was shown essential in improving performance on TED test sets~\cite{zhang-etal-2022-ustc,pham-etal-2022-effective}. 
The end-to-end system, on the other hand, has seen a larger proportion of TED data in training (\autoref{tab:st_data}).

Similar to the previous year \cite{polak-etal-2022-cuni}, we also adapt our end-to-end offline model for simultaneous track \cite{polak-etal-2023-towards}.

\section{Conclusion}
In this paper, we described our systems for the multilingual speech translation track of IWSLT 2023, which translates English speech into $10$ target languages.
To tackle the task of translating scientific conference talks, which feature non-native input speech and terminology-dense contents, our systems have several novelties.
Lacking suitable training data for the target domain, we used $k$NN-MT for inference-time adaptation and showed an improvement of $+0.8$ BLEU for cascaded speech translation system.
We also used adapters to integrate incremental data from augmentation, and achieved performance on-par with re-training on all data.
In our experiments, we observed that cascaded systems are more easily adaptable towards desired target domains due to their separate modules.
Our cascaded speech system outperforms its end-to-end counterpart on scientific talk translation, although their performance remains similar on TED talks.
For future work, we are interested in the feasibility of applying the adaptation approaches shown effective on MT to end-to-end ST.

\paragraph{Acknowledgement}
We thank the anonymous reviewers for detailed and insightful feedback.
Part of this work was performed on the HoreKa supercomputer funded by the Ministry of Science, Research and the Arts Baden-W\"urttemberg and by the Federal Ministry of Education and Research of Germany.
Part of this work was supported by the Federal Ministry of Education and Research of Germany under grant agreement 01EF1803B (RELATER).

\bibliography{anthology,custom}

\begin{thebibliography}{42}
\expandafter\ifx\csname natexlab\endcsname\relax\def\natexlab#1{#1}\fi

\bibitem[{Agarwal et~al.(2023)Agarwal, Agrawal, Anastasopoulos, Bojar, Borg,
  Carpuat, Cattoni, Cettolo, Chen, Chen, Choukri, Chronopoulou, Currey,
  Declerck, Dong, Est\'eve, Duh, Federico, Gahbiche, Haddow, Hsu, Htut,
  Inaguma, Javorsk\'{y}, Judge, Kano, Ko, Kumar, Li, Ma, Mathur, Matusov,
  McNamee, McCrae, Murray, Nadejde, Nakamura, Negri, Nguyen, Niehues, Niu,
  Ojha~Kr., Ortega, Pal, Pino, van~der Plas, Pol\'{a}k, Rippeth, Salesky, Shi,
  Sperber, St\"uker, Sudoh, Tang, Thompson, Tran, Turchi, Waibel, Wang,
  Watanabe, and Zevallos}]{iwslt:2023}
Milind Agarwal, Sweta Agrawal, Antonios Anastasopoulos, Ond\v{r}ej Bojar,
  Claudia Borg, Marine Carpuat, Roldano Cattoni, Mauro Cettolo, Mingda Chen,
  William Chen, Khalid Choukri, Alexandra Chronopoulou, Anna Currey, Thierry
  Declerck, Qianqian Dong, Yannick Est\'eve, Kevin Duh, Marcello Federico,
  Souhir Gahbiche, Barry Haddow, Benjamin Hsu, Phu~Mon Htut, Hirofumi Inaguma,
  D\'{a}vid Javorsk\'{y}, John Judge, Yasumasa Kano, Tom Ko, Rishu Kumar,
  Pengwei Li, Xutai Ma, Prashant Mathur, Evgeny Matusov, Paul McNamee, John~P.
  McCrae, Kenton Murray, Maria Nadejde, Satoshi Nakamura, Matteo Negri,
  Ha~Nguyen, Jan Niehues, Xing Niu, Atul Ojha~Kr., John~E. Ortega, Proyag Pal,
  Juan Pino, Lonneke van~der Plas, Peter Pol\'{a}k, Elijah Rippeth, Elizabeth
  Salesky, Jiatong Shi, Matthias Sperber, Sebastian St\"uker, Katsuhito Sudoh,
  Yun Tang, Brian Thompson, Kevin Tran, Marco Turchi, Alex Waibel, Mingxuan
  Wang, Shinji Watanabe, and Rodolfo Zevallos. 2023.
\newblock Findings of the {IWSLT} 2023 {Evaluation} {Campaign}.
\newblock In \emph{Proceedings of the 20th {International} {Conference} on
  {Spoken} {Language} {Translation} ({IWSLT} 2023)}. Association for
  Computational Linguistics.

\bibitem[{Aharoni et~al.(2019)Aharoni, Johnson, and
  Firat}]{aharoni-etal-2019-massively}
Roee Aharoni, Melvin Johnson, and Orhan Firat. 2019.
\newblock \href {https://doi.org/10.18653/v1/N19-1388} {Massively multilingual
  neural machine translation}.
\newblock In \emph{Proceedings of the 2019 Conference of the North {A}merican
  Chapter of the Association for Computational Linguistics: Human Language
  Technologies, Volume 1 (Long and Short Papers)}, pages 3874--3884,
  Minneapolis, Minnesota. Association for Computational Linguistics.

\bibitem[{Anastasopoulos et~al.(2022)Anastasopoulos, Barrault, Bentivogli,
  Zanon~Boito, Bojar, Cattoni, Currey, Dinu, Duh, Elbayad, Emmanuel,
  Est{\`e}ve, Federico, Federmann, Gahbiche, Gong, Grundkiewicz, Haddow, Hsu,
  Javorsk{\'y}, Kloudov{\'a}, Lakew, Ma, Mathur, McNamee, Murray,
  N{\v{a}}dejde, Nakamura, Negri, Niehues, Niu, Ortega, Pino, Salesky, Shi,
  Sperber, St{\"u}ker, Sudoh, Turchi, Virkar, Waibel, Wang, and
  Watanabe}]{anastasopoulos-etal-2022-findings}
Antonios Anastasopoulos, Lo{\"\i}c Barrault, Luisa Bentivogli, Marcely
  Zanon~Boito, Ond{\v{r}}ej Bojar, Roldano Cattoni, Anna Currey, Georgiana
  Dinu, Kevin Duh, Maha Elbayad, Clara Emmanuel, Yannick Est{\`e}ve, Marcello
  Federico, Christian Federmann, Souhir Gahbiche, Hongyu Gong, Roman
  Grundkiewicz, Barry Haddow, Benjamin Hsu, D{\'a}vid Javorsk{\'y}, V{\u{e}}ra
  Kloudov{\'a}, Surafel Lakew, Xutai Ma, Prashant Mathur, Paul McNamee, Kenton
  Murray, Maria N{\v{a}}dejde, Satoshi Nakamura, Matteo Negri, Jan Niehues,
  Xing Niu, John Ortega, Juan Pino, Elizabeth Salesky, Jiatong Shi, Matthias
  Sperber, Sebastian St{\"u}ker, Katsuhito Sudoh, Marco Turchi, Yogesh Virkar,
  Alexander Waibel, Changhan Wang, and Shinji Watanabe. 2022.
\newblock \href {https://doi.org/10.18653/v1/2022.iwslt-1.10} {Findings of the
  {IWSLT} 2022 evaluation campaign}.
\newblock In \emph{Proceedings of the 19th International Conference on Spoken
  Language Translation (IWSLT 2022)}, pages 98--157, Dublin, Ireland (in-person
  and online). Association for Computational Linguistics.

\bibitem[{Anastasopoulos et~al.(2021)Anastasopoulos, Bojar, Bremerman, Cattoni,
  Elbayad, Federico, Ma, Nakamura, Negri, Niehues, Pino, Salesky, St{\"u}ker,
  Sudoh, Turchi, Waibel, Wang, and Wiesner}]{anastasopoulos-etal-2021-findings}
Antonios Anastasopoulos, Ond{\v{r}}ej Bojar, Jacob Bremerman, Roldano Cattoni,
  Maha Elbayad, Marcello Federico, Xutai Ma, Satoshi Nakamura, Matteo Negri,
  Jan Niehues, Juan Pino, Elizabeth Salesky, Sebastian St{\"u}ker, Katsuhito
  Sudoh, Marco Turchi, Alexander Waibel, Changhan Wang, and Matthew Wiesner.
  2021.
\newblock \href {https://doi.org/10.18653/v1/2021.iwslt-1.1} {{FINDINGS} {OF}
  {THE} {IWSLT} 2021 {EVALUATION} {CAMPAIGN}}.
\newblock In \emph{Proceedings of the 18th International Conference on Spoken
  Language Translation (IWSLT 2021)}, pages 1--29, Bangkok, Thailand (online).
  Association for Computational Linguistics.

\bibitem[{Ardila et~al.(2020)Ardila, Branson, Davis, Kohler, Meyer, Henretty,
  Morais, Saunders, Tyers, and Weber}]{cv}
Rosana Ardila, Megan Branson, Kelly Davis, Michael Kohler, Josh Meyer, Michael
  Henretty, Reuben Morais, Lindsay Saunders, Francis~M. Tyers, and Gregor
  Weber. 2020.
\newblock \href {https://aclanthology.org/2020.lrec-1.520/} {Common voice: {A}
  massively-multilingual speech corpus}.
\newblock In \emph{Proceedings of The 12th Language Resources and Evaluation
  Conference, {LREC} 2020, Marseille, France, May 11-16, 2020}, pages
  4218--4222. European Language Resources Association.

\bibitem[{Arivazhagan et~al.(2019)Arivazhagan, Bapna, Firat, Lepikhin, Johnson,
  Krikun, Chen, Cao, Foster, Cherry, Macherey, Chen, and
  Wu}]{arivazhagan2019massively}
Naveen Arivazhagan, Ankur Bapna, Orhan Firat, Dmitry Lepikhin, Melvin Johnson,
  Maxim Krikun, Mia~Xu Chen, Yuan Cao, George~F. Foster, Colin Cherry, Wolfgang
  Macherey, Zhifeng Chen, and Yonghui Wu. 2019.
\newblock \href {http://arxiv.org/abs/1907.05019} {Massively multilingual
  neural machine translation in the wild: Findings and challenges}.
\newblock \emph{CoRR}, abs/1907.05019.

\bibitem[{Baevski et~al.(2020)Baevski, Zhou, Mohamed, and Auli}]{w2v2}
Alexei Baevski, Yuhao Zhou, Abdelrahman Mohamed, and Michael Auli. 2020.
\newblock \href
  {https://proceedings.neurips.cc/paper/2020/hash/92d1e1eb1cd6f9fba3227870bb6d7f07-Abstract.html}
  {wav2vec 2.0: {A} framework for self-supervised learning of speech
  representations}.
\newblock In \emph{Advances in Neural Information Processing Systems 33: Annual
  Conference on Neural Information Processing Systems 2020, NeurIPS 2020,
  December 6-12, 2020, virtual}.

\bibitem[{Bapna and Firat(2019)}]{bapna-firat-2019-simple}
Ankur Bapna and Orhan Firat. 2019.
\newblock \href {https://doi.org/10.18653/v1/D19-1165} {Simple, scalable
  adaptation for neural machine translation}.
\newblock In \emph{Proceedings of the 2019 Conference on Empirical Methods in
  Natural Language Processing and the 9th International Joint Conference on
  Natural Language Processing (EMNLP-IJCNLP)}, pages 1538--1548, Hong Kong,
  China. Association for Computational Linguistics.

\bibitem[{Baziotis et~al.(2022)Baziotis, Artetxe, Cross, and
  Bhosale}]{baziotis-etal-2022-multilingual}
Christos Baziotis, Mikel Artetxe, James Cross, and Shruti Bhosale. 2022.
\newblock \href {https://aclanthology.org/2022.emnlp-main.77} {Multilingual
  machine translation with hyper-adapters}.
\newblock In \emph{Proceedings of the 2022 Conference on Empirical Methods in
  Natural Language Processing}, pages 1170--1185, Abu Dhabi, United Arab
  Emirates. Association for Computational Linguistics.

\bibitem[{Chen et~al.(2022)Chen, Wang, Chen, Wu, Liu, Chen, Li, Kanda,
  Yoshioka, Xiao, Wu, Zhou, Ren, Qian, Qian, Wu, Zeng, Yu, and Wei}]{wavlm}
Sanyuan Chen, Chengyi Wang, Zhengyang Chen, Yu~Wu, Shujie Liu, Zhuo Chen, Jinyu
  Li, Naoyuki Kanda, Takuya Yoshioka, Xiong Xiao, Jian Wu, Long Zhou, Shuo Ren,
  Yanmin Qian, Yao Qian, Jian Wu, Michael Zeng, Xiangzhan Yu, and Furu Wei.
  2022.
\newblock \href {https://doi.org/10.1109/JSTSP.2022.3188113} {Wavlm:
  Large-scale self-supervised pre-training for full stack speech processing}.
\newblock \emph{{IEEE} J. Sel. Top. Signal Process.}, 16(6):1505--1518.

\bibitem[{Di~Gangi et~al.(2019)Di~Gangi, Cattoni, Bentivogli, Negri, and
  Turchi}]{di-gangi-etal-2019-must}
Mattia~A. Di~Gangi, Roldano Cattoni, Luisa Bentivogli, Matteo Negri, and Marco
  Turchi. 2019.
\newblock \href {https://doi.org/10.18653/v1/N19-1202} {{M}u{ST}-{C}: a
  {M}ultilingual {S}peech {T}ranslation {C}orpus}.
\newblock In \emph{Proceedings of the 2019 Conference of the North {A}merican
  Chapter of the Association for Computational Linguistics: Human Language
  Technologies, Volume 1 (Long and Short Papers)}, pages 2012--2017,
  Minneapolis, Minnesota. Association for Computational Linguistics.

\bibitem[{Eck et~al.(2005)Eck, Vogel, and Waibel}]{eck-etal-2005-low}
Matthias Eck, Stephan Vogel, and Alex Waibel. 2005.
\newblock \href {https://aclanthology.org/2005.iwslt-1.7} {Low cost portability
  for statistical machine translation based on n-gram frequency and
  {TF}-{IDF}}.
\newblock In \emph{Proceedings of the Second International Workshop on Spoken
  Language Translation}, Pittsburgh, Pennsylvania, USA.

\bibitem[{Fan et~al.(2021)Fan, Bhosale, Schwenk, Ma, El{-}Kishky, Goyal,
  Baines, Celebi, Wenzek, Chaudhary, Goyal, Birch, Liptchinsky, Edunov, Auli,
  and Joulin}]{mm100}
Angela Fan, Shruti Bhosale, Holger Schwenk, Zhiyi Ma, Ahmed El{-}Kishky,
  Siddharth Goyal, Mandeep Baines, Onur Celebi, Guillaume Wenzek, Vishrav
  Chaudhary, Naman Goyal, Tom Birch, Vitaliy Liptchinsky, Sergey Edunov,
  Michael Auli, and Armand Joulin. 2021.
\newblock \href {http://jmlr.org/papers/v22/20-1307.html} {Beyond
  english-centric multilingual machine translation}.
\newblock \emph{The Journal of Machine Learning Research}, 22:107:1--107:48.

\bibitem[{Hernandez et~al.(2018)Hernandez, Nguyen, Ghannay, Tomashenko, and
  Est{\`{e}}ve}]{tedlium3}
Fran{\c{c}}ois Hernandez, Vincent Nguyen, Sahar Ghannay, Natalia~A. Tomashenko,
  and Yannick Est{\`{e}}ve. 2018.
\newblock \href {https://doi.org/10.1007/978-3-319-99579-3\_21} {{TED-LIUM} 3:
  Twice as much data and corpus repartition for experiments on speaker
  adaptation}.
\newblock In \emph{Speech and Computer - 20th International Conference,
  {SPECOM} 2018, Leipzig, Germany, September 18-22, 2018, Proceedings}, volume
  11096 of \emph{Lecture Notes in Computer Science}, pages 198--208. Springer.

\bibitem[{Iranzo{-}S{\'{a}}nchez et~al.(2020)Iranzo{-}S{\'{a}}nchez,
  Silvestre{-}Cerd{\`{a}}, Jorge, Rosell{\'{o}}, Gim{\'{e}}nez, Sanch{\'{\i}}s,
  Civera, and Juan}]{europarlst}
Javier Iranzo{-}S{\'{a}}nchez, Joan~Albert Silvestre{-}Cerd{\`{a}}, Javier
  Jorge, Nahuel Rosell{\'{o}}, Adri{\`{a}} Gim{\'{e}}nez, Albert
  Sanch{\'{\i}}s, Jorge Civera, and Alfons Juan. 2020.
\newblock \href {https://doi.org/10.1109/ICASSP40776.2020.9054626}
  {Europarl-st: {A} multilingual corpus for speech translation of parliamentary
  debates}.
\newblock In \emph{2020 {IEEE} International Conference on Acoustics, Speech
  and Signal Processing, {ICASSP} 2020, Barcelona, Spain, May 4-8, 2020}, pages
  8229--8233. {IEEE}.

\bibitem[{Johnson et~al.(2019)Johnson, Douze, and
  J{\'e}gou}]{johnson2019billion}
Jeff Johnson, Matthijs Douze, and Herv{\'e} J{\'e}gou. 2019.
\newblock Billion-scale similarity search with {GPUs}.
\newblock \emph{IEEE Transactions on Big Data}, 7(3):535--547.

\bibitem[{Khandelwal et~al.(2021)Khandelwal, Fan, Jurafsky, Zettlemoyer, and
  Lewis}]{knnmt}
Urvashi Khandelwal, Angela Fan, Dan Jurafsky, Luke Zettlemoyer, and Mike Lewis.
  2021.
\newblock \href {https://openreview.net/forum?id=7wCBOfJ8hJM} {Nearest neighbor
  machine translation}.
\newblock In \emph{9th International Conference on Learning Representations,
  {ICLR} 2021, Virtual Event, Austria, May 3-7, 2021}. OpenReview.net.

\bibitem[{Kim et~al.(2021)Kim, Kong, and Son}]{vits}
Jaehyeon Kim, Jungil Kong, and Juhee Son. 2021.
\newblock \href {http://proceedings.mlr.press/v139/kim21f.html} {Conditional
  variational autoencoder with adversarial learning for end-to-end
  text-to-speech}.
\newblock In \emph{Proceedings of the 38th International Conference on Machine
  Learning, {ICML} 2021, 18-24 July 2021, Virtual Event}, volume 139 of
  \emph{Proceedings of Machine Learning Research}, pages 5530--5540. {PMLR}.

\bibitem[{Koehn(2005)}]{koehn-2005-europarl}
Philipp Koehn. 2005.
\newblock \href {https://aclanthology.org/2005.mtsummit-papers.11} {{E}uroparl:
  A parallel corpus for statistical machine translation}.
\newblock In \emph{Proceedings of Machine Translation Summit X: Papers}, pages
  79--86, Phuket, Thailand.

\bibitem[{Koneru et~al.(2022)Koneru, Liu, and
  Niehues}]{DBLP:journals/corr/abs-2201-05700}
Sai Koneru, Danni Liu, and Jan Niehues. 2022.
\newblock \href {http://arxiv.org/abs/2201.05700} {Cost-effective training in
  low-resource neural machine translation}.
\newblock \emph{CoRR}, abs/2201.05700.

\bibitem[{Kudo and Richardson(2018)}]{kudo-richardson-2018-sentencepiece}
Taku Kudo and John Richardson. 2018.
\newblock \href {https://doi.org/10.18653/v1/D18-2012} {{S}entence{P}iece: A
  simple and language independent subword tokenizer and detokenizer for neural
  text processing}.
\newblock In \emph{Proceedings of the 2018 Conference on Empirical Methods in
  Natural Language Processing: System Demonstrations}, pages 66--71, Brussels,
  Belgium. Association for Computational Linguistics.

\bibitem[{Lison and Tiedemann(2016)}]{lison-tiedemann-2016-opensubtitles2016}
Pierre Lison and J{\"o}rg Tiedemann. 2016.
\newblock \href {https://aclanthology.org/L16-1147} {{O}pen{S}ubtitles2016:
  Extracting large parallel corpora from movie and {TV} subtitles}.
\newblock In \emph{Proceedings of the Tenth International Conference on
  Language Resources and Evaluation ({LREC}'16)}, pages 923--929,
  Portoro{\v{z}}, Slovenia. European Language Resources Association (ELRA).

\bibitem[{Ma et~al.(2021)Ma, Dong, Huang, Zhang, Muzio, Singhal, Awadalla,
  Song, and Wei}]{deltalm}
Shuming Ma, Li~Dong, Shaohan Huang, Dongdong Zhang, Alexandre Muzio, Saksham
  Singhal, Hany~Hassan Awadalla, Xia Song, and Furu Wei. 2021.
\newblock \href {http://arxiv.org/abs/2106.13736} {Deltalm: Encoder-decoder
  pre-training for language generation and translation by augmenting pretrained
  multilingual encoders}.
\newblock \emph{CoRR}, abs/2106.13736.

\bibitem[{Morishita et~al.(2022)Morishita, Chousa, Suzuki, and
  Nagata}]{morishita-etal-2022-jparacrawl}
Makoto Morishita, Katsuki Chousa, Jun Suzuki, and Masaaki Nagata. 2022.
\newblock \href {https://aclanthology.org/2022.lrec-1.721} {{JP}ara{C}rawl
  v3.0: A large-scale {E}nglish-{J}apanese parallel corpus}.
\newblock In \emph{Proceedings of the Thirteenth Language Resources and
  Evaluation Conference}, pages 6704--6710, Marseille, France. European
  Language Resources Association.

\bibitem[{Nguyen and Salazar(2019)}]{prenorm}
Toan~Q. Nguyen and Julian Salazar. 2019.
\newblock \href {https://aclanthology.org/2019.iwslt-1.17} {Transformers
  without tears: Improving the normalization of self-attention}.
\newblock In \emph{Proceedings of the 16th International Conference on Spoken
  Language Translation, {IWSLT} 2019, Hong Kong, November 2-3, 2019}.
  Association for Computational Linguistics.

\bibitem[{Nguyen et~al.(2020)Nguyen, Joty, Wu, and Aw}]{data_div}
Xuan{-}Phi Nguyen, Shafiq~R. Joty, Kui Wu, and Ai~Ti Aw. 2020.
\newblock \href
  {https://proceedings.neurips.cc/paper/2020/hash/7221e5c8ec6b08ef6d3f9ff3ce6eb1d1-Abstract.html}
  {Data diversification: {A} simple strategy for neural machine translation}.
\newblock In \emph{Advances in Neural Information Processing Systems 33: Annual
  Conference on Neural Information Processing Systems 2020, NeurIPS 2020,
  December 6-12, 2020, virtual}.

\bibitem[{Panayotov et~al.(2015)Panayotov, Chen, Povey, and
  Khudanpur}]{librispeech}
Vassil Panayotov, Guoguo Chen, Daniel Povey, and Sanjeev Khudanpur. 2015.
\newblock \href {https://doi.org/10.1109/ICASSP.2015.7178964} {Librispeech: An
  {ASR} corpus based on public domain audio books}.
\newblock In \emph{2015 {IEEE} International Conference on Acoustics, Speech
  and Signal Processing, {ICASSP} 2015, South Brisbane, Queensland, Australia,
  April 19-24, 2015}, pages 5206--5210. {IEEE}.

\bibitem[{Pham et~al.(2022)Pham, Nguyen, Nguyen, Liu, Mullov, Niehues, and
  Waibel}]{pham-etal-2022-effective}
Ngoc-Quan Pham, Tuan~Nam Nguyen, Thai-Binh Nguyen, Danni Liu, Carlos Mullov,
  Jan Niehues, and Alexander Waibel. 2022.
\newblock \href {https://doi.org/10.18653/v1/2022.iwslt-1.14} {Effective
  combination of pretrained models - {KIT}@{IWSLT}2022}.
\newblock In \emph{Proceedings of the 19th International Conference on Spoken
  Language Translation (IWSLT 2022)}, pages 190--197, Dublin, Ireland
  (in-person and online). Association for Computational Linguistics.

\bibitem[{Philip et~al.(2020)Philip, Berard, Gall{\'e}, and
  Besacier}]{philip-etal-2020-monolingual}
Jerin Philip, Alexandre Berard, Matthias Gall{\'e}, and Laurent Besacier. 2020.
\newblock \href {https://doi.org/10.18653/v1/2020.emnlp-main.361} {Monolingual
  adapters for zero-shot neural machine translation}.
\newblock In \emph{Proceedings of the 2020 Conference on Empirical Methods in
  Natural Language Processing (EMNLP)}, pages 4465--4470, Online. Association
  for Computational Linguistics.

\bibitem[{Pires et~al.(2023)Pires, Schmidt, Liao, and
  Peitz}]{DBLP:journals/corr/abs-2305-02665}
Telmo~Pessoa Pires, Robin~M. Schmidt, Yi{-}Hsiu Liao, and Stephan Peitz. 2023.
\newblock \href {https://doi.org/10.48550/arXiv.2305.02665} {Learning
  language-specific layers for multilingual machine translation}.
\newblock \emph{CoRR}, abs/2305.02665.

\bibitem[{Pol{\'a}k et~al.(2023)Pol{\'a}k, Liu, Pham, Niehues, Waibel, and
  Bojar}]{polak-etal-2023-towards}
Peter Pol{\'a}k, Danni Liu, Ngoc-Quan Pham, Jan Niehues, Alexander Waibel, and
  Ond{\v{r}}ej Bojar. 2023.
\newblock Towards efficient simultaneous speech translation: {CUNI}-{KIT}
  system for simultaneous track at {IWSLT} 2023.
\newblock In \emph{Proceedings of the 20th International Conference on Spoken
  Language Translation (IWSLT 2023)}, Toronto, Canada (in-person and online).
  Association for Computational Linguistics.

\bibitem[{Pol{\'a}k et~al.(2022)Pol{\'a}k, Pham, Nguyen, Liu, Mullov, Niehues,
  Bojar, and Waibel}]{polak-etal-2022-cuni}
Peter Pol{\'a}k, Ngoc-Quan Pham, Tuan~Nam Nguyen, Danni Liu, Carlos Mullov, Jan
  Niehues, Ond{\v{r}}ej Bojar, and Alexander Waibel. 2022.
\newblock \href {https://doi.org/10.18653/v1/2022.iwslt-1.24} {{CUNI}-{KIT}
  system for simultaneous speech translation task at {IWSLT} 2022}.
\newblock In \emph{Proceedings of the 19th International Conference on Spoken
  Language Translation (IWSLT 2022)}, pages 277--285, Dublin, Ireland
  (in-person and online). Association for Computational Linguistics.

\bibitem[{Post(2018)}]{post-2018-call}
Matt Post. 2018.
\newblock \href {https://doi.org/10.18653/v1/W18-6319} {A call for clarity in
  reporting {BLEU} scores}.
\newblock In \emph{Proceedings of the Third Conference on Machine Translation:
  Research Papers}, pages 186--191, Brussels, Belgium. Association for
  Computational Linguistics.

\bibitem[{Rebuffi et~al.(2017)Rebuffi, Bilen, and Vedaldi}]{adapters_orig}
Sylvestre{-}Alvise Rebuffi, Hakan Bilen, and Andrea Vedaldi. 2017.
\newblock \href
  {https://proceedings.neurips.cc/paper/2017/hash/e7b24b112a44fdd9ee93bdf998c6ca0e-Abstract.html}
  {Learning multiple visual domains with residual adapters}.
\newblock In \emph{Advances in Neural Information Processing Systems 30: Annual
  Conference on Neural Information Processing Systems 2017, December 4-9, 2017,
  Long Beach, CA, {USA}}, pages 506--516.

\bibitem[{Reimers and Gurevych(2020)}]{reimers-gurevych-2020-making}
Nils Reimers and Iryna Gurevych. 2020.
\newblock \href {https://doi.org/10.18653/v1/2020.emnlp-main.365} {Making
  monolingual sentence embeddings multilingual using knowledge distillation}.
\newblock In \emph{Proceedings of the 2020 Conference on Empirical Methods in
  Natural Language Processing (EMNLP)}, pages 4512--4525, Online. Association
  for Computational Linguistics.

\bibitem[{Salesky et~al.(2023)Salesky, Darwish, Al-Badrashiny, Diab, and
  Niehues}]{salesky-2023-evaluating}
Elizabeth Salesky, Kareem Darwish, Mohamed Al-Badrashiny, Mona Diab, and Jan
  Niehues. 2023.
\newblock {Evaluating Multilingual Speech Translation Under Realistic
  Conditions with Resegmentation and Terminology}.
\newblock In \emph{Proceedings of the 20th International Conference on Spoken
  Language Translation (IWSLT 2023)}. Association for Computational
  Linguistics.

\bibitem[{Tang et~al.(2020)Tang, Tran, Li, Chen, Goyal, Chaudhary, Gu, and
  Fan}]{mbart50}
Yuqing Tang, Chau Tran, Xian Li, Peng{-}Jen Chen, Naman Goyal, Vishrav
  Chaudhary, Jiatao Gu, and Angela Fan. 2020.
\newblock \href {http://arxiv.org/abs/2008.00401} {Multilingual translation
  with extensible multilingual pretraining and finetuning}.
\newblock \emph{CoRR}, abs/2008.00401.

\bibitem[{Tiedemann(2012)}]{tiedemann-2012-parallel}
J{\"o}rg Tiedemann. 2012.
\newblock \href
  {http://www.lrec-conf.org/proceedings/lrec2012/pdf/463_Paper.pdf} {Parallel
  data, tools and interfaces in {OPUS}}.
\newblock In \emph{Proceedings of the Eighth International Conference on
  Language Resources and Evaluation ({LREC}'12)}, pages 2214--2218, Istanbul,
  Turkey. European Language Resources Association (ELRA).

\bibitem[{Wang et~al.(2021)Wang, Riviere, Lee, Wu, Talnikar, Haziza,
  Williamson, Pino, and Dupoux}]{wang-etal-2021-voxpopuli}
Changhan Wang, Morgane Riviere, Ann Lee, Anne Wu, Chaitanya Talnikar, Daniel
  Haziza, Mary Williamson, Juan Pino, and Emmanuel Dupoux. 2021.
\newblock \href {https://doi.org/10.18653/v1/2021.acl-long.80} {{V}ox{P}opuli:
  A large-scale multilingual speech corpus for representation learning,
  semi-supervised learning and interpretation}.
\newblock In \emph{Proceedings of the 59th Annual Meeting of the Association
  for Computational Linguistics and the 11th International Joint Conference on
  Natural Language Processing (Volume 1: Long Papers)}, pages 993--1003,
  Online. Association for Computational Linguistics.

\bibitem[{Wang et~al.(2020)Wang, Wu, and Pino}]{covost}
Changhan Wang, Anne Wu, and Juan~Miguel Pino. 2020.
\newblock \href {http://arxiv.org/abs/2007.10310} {Covost 2: {A} massively
  multilingual speech-to-text translation corpus}.
\newblock \emph{CoRR}, abs/2007.10310.

\bibitem[{Zhang et~al.(2022)Zhang, Ye, Tang, Li, Zhou, Yang, Cui, Deng, Shi,
  Song, Liu, Liu, and Dai}]{zhang-etal-2022-ustc}
Weitai Zhang, Zhongyi Ye, Haitao Tang, Xiaoxi Li, Xinyuan Zhou, Jing Yang,
  Jianwei Cui, Pan Deng, Mohan Shi, Yifan Song, Dan Liu, Junhua Liu, and Lirong
  Dai. 2022.
\newblock \href {https://doi.org/10.18653/v1/2022.iwslt-1.15} {The
  {USTC}-{NELSLIP} offline speech translation systems for {IWSLT} 2022}.
\newblock In \emph{Proceedings of the 19th International Conference on Spoken
  Language Translation (IWSLT 2022)}, pages 198--207, Dublin, Ireland
  (in-person and online). Association for Computational Linguistics.

\bibitem[{Zheng et~al.(2021)Zheng, Zhang, Guo, Huang, Chen, Luo, and
  Chen}]{zheng-etal-2021-adaptive}
Xin Zheng, Zhirui Zhang, Junliang Guo, Shujian Huang, Boxing Chen, Weihua Luo,
  and Jiajun Chen. 2021.
\newblock \href {https://doi.org/10.18653/v1/2021.acl-short.47} {Adaptive
  nearest neighbor machine translation}.
\newblock In \emph{Proceedings of the 59th Annual Meeting of the Association
  for Computational Linguistics and the 11th International Joint Conference on
  Natural Language Processing (Volume 2: Short Papers)}, pages 368--374,
  Online. Association for Computational Linguistics.

\end{thebibliography}

% \appendix

% \section{Example Appendix}
% \label{sec:appendix}

% This is an appendix.

\end{document}